\documentclass[a4paper,conference]{IEEEtran}

\ifCLASSINFOpdf
\else
\fi
\usepackage{graphicx}
\usepackage{cite}
\usepackage{amsmath,amssymb,amsfonts}
\usepackage{algorithmic}
\usepackage{textcomp}
\usepackage{xcolor}
\usepackage{graphicx}
\usepackage{caption}
\usepackage{float}
\usepackage{subcaption}
\usepackage{booktabs}
\usepackage{multirow}
\usepackage{comment}
\usepackage{lipsum}
\newcommand\blfootnote[1]{%
  \begingroup
  \renewcommand\thefootnote{}\footnote{#1}%
  \addtocounter{footnote}{-1}%
  \endgroup
}
\newcommand{\RomanNumeralCaps}[1]
{\MakeUppercase{\romannumeral #1}}
\begin{document}
%
\title{On-Device Text Image Super Resolution}

\author{\IEEEauthorblockN{Dhruval Jain*}
\IEEEauthorblockA{OnDevice AI\\
Samsung R\& D Institute\\
Bangalore, India\\
dhruval.jain@samsung.com}
\and
\IEEEauthorblockN{Arun D Prabhu*}
\IEEEauthorblockA{OnDevice AI\\
Samsung R\& D Institute\\
Bangalore, India\\
arun.prabhu@samsung.com}
\and
\IEEEauthorblockN{Gopi Ramena}
\IEEEauthorblockA{OnDevice AI\\
Samsung R\& D Institute\\
Bangalore, India\\
gopi.ramena@samsung.com}
\and
\IEEEauthorblockN{Manoj Goyal}
\IEEEauthorblockA{OnDevice AI\\
Samsung R\& D Institute\\
Bangalore, India\\
manoj.goyal@samsung.com}
\and
\hspace{2cm}
\IEEEauthorblockN{Debi Prasanna Mohanty}
\IEEEauthorblockA{\hspace{2cm} OnDevice AI\\
\hspace{2cm}Samsung R\& D Institute\\
\hspace{2cm}Bangalore, India\\
\hspace{2cm}debi.m@samsung.com}
\and
\IEEEauthorblockN{Sukumar Moharana}
\IEEEauthorblockA{OnDevice AI\\
Samsung R\& D Institute\\
Bangalore, India\\
msukumar@samsung.com}
\and
\IEEEauthorblockN{Naresh Purre}
\IEEEauthorblockA{OnDevice AI\\
Samsung R\& D Institute\\
Bangalore, India\\
naresh.purre@samsung.com}
}


\maketitle

\begin{abstract}
Recent\blfootnote{* Primary authors} research on super-resolution (SR) has witnessed major developments with the advancements of deep convolutional neural networks. There is a need for information extraction from scenic text images or even document images on device, most of which are low-resolution (LR) images. Therefore, SR becomes an essential pre-processing step as Bicubic Upsampling, which is conventionally present in smartphones, performs poorly on LR images. To give the user more control over his privacy, and to reduce the carbon footprint by reducing the overhead of cloud computing and hours of GPU usage, executing SR models on the edge is a necessity in the recent times. There are various challenges in running and optimizing a model on resource-constrained platforms like smartphones. In this paper, we present a novel deep neural network that reconstructs sharper character edges and thus boosts OCR confidence. The proposed architecture not only achieves significant improvement in PSNR over bicubic upsampling on various benchmark datasets but also runs with an average inference time of 11.7 ms per image. We have outperformed state-of-the-art on the Text330 dataset. We also achieve an OCR accuracy of 75.89\% on the ICDAR 2015 TextSR dataset, where ground truth has an accuracy of 78.10\%.

\end{abstract}


%
\IEEEpeerreviewmaketitle

\section{Introduction}

Compression techniques used by social media platforms significantly reduce the resolution of images adversely affecting the finer details. Due to the large amounts of images being shared over social media today, there is a growing abundance of low resolution (LR) images on smartphones. LR text images thus, pose a challenge for existing text detection and recognition frameworks on device. Moreover, it gets difficult for the user to read the text due to blurry character edges. Hence, super-resolution (SR) of text images is an intuitive solution. SR is a well-studied problem in computer vision and various approaches have been proposed in the literature. 

Some very early approaches use interpolation techniques \cite{allebach1996edge, li2001new, zhang2006edge} based on sampling theory. Many previous studies even adopted edge-based \cite{sun2010gradient} techniques and patch-based techniques \cite{chang2004super, gao2012image, zhu2014single} to approach the problem. Another study \cite{yang2008image} showed good results using sparsity-based methods. Sparse coding assumes any image can be sparsely represented in a transform domain, which can then be learned through a training mechanism that tries to relate the LR and HR images. Yang et al. \cite{yang2013fast, yang2012coupled} introduced another approach by clustering the patch spaces and learning the corresponding functions. Dong et al. \cite{dong2014learning, dong2016accelerating} proposed SRCNN, the first deep learning-based image SR method. Since then, various CNN model architectures have been studied for image SR in recent years. Extending this further, other studies \cite{vdsr, drcn} have also used Residual neural networks\cite{resnet} for achieving better performances. They employ deeper neural network architectures and show that using skip connections and recursive convolutions improve SR networks by carrying forward the identity information ahead in the network. Mao et al. \cite{mao2016image} used encoder-decoder networks and symmetric skip connections to tackle the problem as image restoration. They show that the skip connections help the model in converging faster. Stacked collaborative autoencoders are also studied in \cite{cui2014deep}. Generative Adversarial Networks (GANs)\cite{goodfellow2014generative} introduce adversarial loss to generate impressive results in various tasks. SRGAN \cite{srgan} is the first work to apply GANs to image SR.

The above mentioned SR frameworks have focussed on learning the texture present in the LR image and reproducing it in the predicted HR with refined details. So, they leverage deeper neural networks to learn hierarchical features for preserving the texture. Reconstruction of texture details is of lesser importance in improving OCR confidence than learning sharper character outline edges.

In this paper, we propose a novel text image SR solution that runs on device in real-time. Our proposed network with very less number of parameters produces comparable results with existing state-of-the-art networks on ICDAR 2015 TextSR
\footnote{https://projet.liris.cnrs.fr/sr2015/} datasets in terms of OCR accuracy. The feature extraction pipeline pools high-level textual features for finner edge reconstruction. Our model learns upon the bicubic interpolation hence reducing model parameters to a significant extent. It shows significant improvement in text clarity compared to bicubic interpolation on device.

\section{Related Work}

Earliest works in text image super-resolution include \cite{capel} and \cite{nasonov}. Capel et al.\cite{nasonov} proposed the enhancement of text images using multiple-frame SR. They introduced a MAP estimator based on Huber prior and utilized Total Variation for regularization. Extending their work, Nasonov\cite{nasonov} combined Bilateral Total Variation\cite{btv} and a penalty function based on bimodal prior that forces the pixel intensities of the reconstructed image to be either close to the background or the foreground (text). However, these methods need many LR observations for effective estimation and were only restricted to 2$\times$ upscaling. Banerjee et al.\cite{banerjee} utilized the tangent field at character edges for reconstructing sharper edges in grayscale document LR images. A Lat et al.\cite{lat} improved OCR accuracy by super-resolving document images using SRGAN\cite{srgan}. Another closely related work, \cite{pandey2018binary} uses deep neural network architectures for Binary Document Image Super-Resolution (BDISR) in Tamil script. These works are only restricted to document images, whereas our method solves a broader problem of text image SR while improving OCR confidence. \cite{srgan} produces HR image with higher fidelity incorporating finer edge details but has used GANs which are not suitable for a smartphone environment. \cite{sredgenet} has proposed MergeNet for enhancing the edges in the HR output from their version of EDSR\cite{edsr}. Their pipelined architecture has huge memory constraints and can not be run in real-time on device.

C Dong et al. in \cite{dong} had trained 11 different variants of SRCNN on the ICDAR2015 TextSR dataset. They used Model combination\cite{combination}, following "Greedy search strategy\cite{dong}" to obtain 14-best model combinations evaluated on 
peak signal-to-noise ratio (PSNR). The 14-model combination achieves state-of-the-art performance on the test set. More recent works, driven by the powerful capability of deep neural networks, has led to dramatic improvements in SR. Other works in text image SR\cite{hanh,haochen,weighted}, have trained deeper neural networks that have shown state-of-the-art performance in image super-resolution like LapSRN\cite{lapsrn}, VDSR\cite{vdsr} with loss functions tailored for reconstructing sharper letter edges. However, these solutions can not be deployed on the device due to their large number of parameters.

MobiSR\cite{mobisr} introduced hardware optimization to perform efficient image SR on mobile platforms by minimizing processing latency. Their novel scheduler estimates the upscaling difficulty patch-wise and accordingly sends them to model-engines. Model compression techniques like Knowledge Distillation\cite{kd}, pruning and quantization have been utilized to reduce the inference time of the network. However, they may also result in a significant loss of high-frequency features that further affect the sharpness of reconstructed edges in the HR image. Therefore, we introduce a shallow network that focuses on learning sharper character edges for enhanced readability and faster inference time on device. To the best of our knowledge, we are the first to perform text image SR on device.

\begin{figure*}[htbp]
\includegraphics[width=180mm]{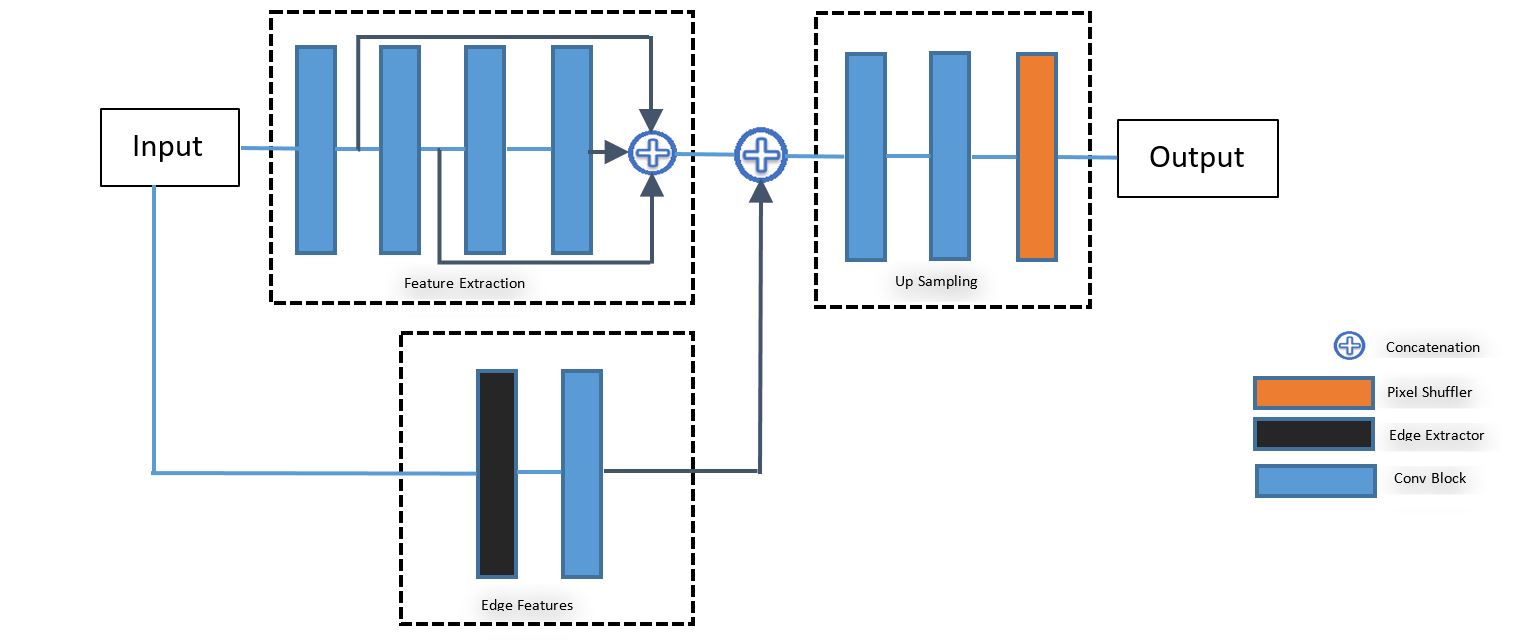}
\label{fig:label}
\centering
\caption{Architecture}
\centering
\end{figure*}

\section{Methodology}

Mean Squared Error (MSE) loss tends to induce smoothness in the reconstructed HR image which results in blurry edges. H Zhang et al.\cite{haochen} have trained VDSR\cite{vdsr} with the proposed edge-based Weighted Mean Squared error (WMSE) loss. The weight at any pixel is the value of some function $f$ of the gradient magnitude at that pixel, where $f$ is any monotonously increasing function. This has shown a slight improvement in OCR confidence on ICDAR 2015 TextSR test set. In our initial experiments, we tried to add edge information using different loss functions \cite{weighted, gdl, haochen}. We found that shallower models are unable to reflect this impact in the reconstructed HR image. Our architecture is based on residual learning and thus, it learns the residual between bicubic interpolated HR and the ground truth. Various components of the model are described in detail as follows:

\subsection{Feature Extraction}
Pre-upsampling SR networks like SRCNN\cite{srcnn}, DRRN\cite{drrn}, DRCN\cite{drcn} extract a set of features from bicubic interpolated LR image. A couple of convolution layers are then leveraged to perform non-linear mapping of the extracted feature maps and reconstruct the HR image. Text images suffer a significant loss in edge information due to blurring and noise amplification resulted from bicubic upsampling (Fig. 2).
Reconstructing HR from the feature maps extracted from the upsampled bicubic text image does not produce expected results. Moreover, it also increases the space and time complexity as the operations are performed in the high dimensional space. So, we adopt the post-upsampling SR approach \cite{edsr,skipconnSR} as shown in the Fig. 1. High frequency detailing like sharper edge transitions between text and the background is crucial for enhanced readability. For faster inference on device, we convert the RGB input to $YC_{b}C_{r}$ space and only Y channel is fed to the network for training.

We use four CNN layers having 32, 24, 16, 8 filters respectively each with kernel size 3$\times$3 to extract features from the LR input image. All the feature maps except the third layer are concatenated via skip connections to generate a total of 64 feature maps. Though, skip connections were introduced to eliminate the vanishing gradient problem in very deep neural networks. \cite{skipconnSR} has shown improved results over SRCNN, VDSR, and DRCN using dense skip connections. \cite{dcscn} also used skip connections in their feature extraction pipeline to extract both local and global features. Decreasing order of filters and skip connections from the outputs of initial layers ensure that high-frequency features are well incorporated. Our idea behind pooling features from initial layers is to add more emphasis on broad edges that outline the character boundaries in text images. Our shallow model has only \textbf{57K} parameters.

Additionally, we have also extracted features from Sobel edge maps of the input LR 
using a convolution layer having 16 filters and a kernel size of 3$\times$3 as shown in Fig. 1. Edge feature maps are then concatenated with the existing ones. This helps in emphasizing weaker edges and in turn, fills the fine broken edges in the reconstructed HR edge map. It has also led to a slight improvement in metrics as shown in TABLE \RomanNumeralCaps{2}.

\subsection{Upsampling Layer}
The generated feature maps are further passed to two convolution layers having 32 and $s^{2}$ filters respectively, where s is the scale of resolution. Assuming input images shape is h $\times$ w, the resultant input to the pixel shuffler layer is h $\times$ w $\times$ $s^{2}$. The reshaping operation is performed through pixel shuffle \cite{checkerboard} generates the final HR image of the Y-channel input LR having the shape hs $\times$ ws.   

\begin{figure*}[t!]
	
	\hspace*{\fill}
    \begin{subfigure}[b]{0.32\textwidth}
        \includegraphics[width=\textwidth]{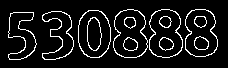}
    \end{subfigure}
    \hspace*{\fill}
    \begin{subfigure}[b]{0.32\textwidth}
        \includegraphics[width=\textwidth]{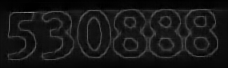}
    \end{subfigure}
    \hspace*{\fill}
	\begin{subfigure}[b]{0.32\textwidth}
        \includegraphics[width=\textwidth]{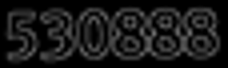}
    \end{subfigure}
    \hspace*{\fill} \\

    \hspace*{\fill}
    \begin{subfigure}[b]{0.32\textwidth}
        \includegraphics[width=\textwidth]{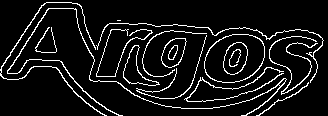}
    \end{subfigure}
    \hspace*{\fill}
    \begin{subfigure}[b]{0.32\textwidth}
        \includegraphics[width=\textwidth]{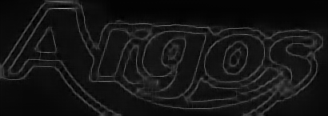}
    \end{subfigure}
    \hspace*{\fill}
	\begin{subfigure}[b]{0.32\textwidth}
        \includegraphics[width=\textwidth]{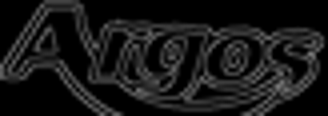}
    \end{subfigure}
    \hspace*{\fill} \\
   
    \hspace*{\fill}
    \begin{subfigure}[b]{0.32\textwidth}
        \includegraphics[width=\textwidth]{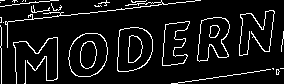}
        \caption{Input Canny Edge Map}
    \end{subfigure}
    \hspace*{\fill}
    \begin{subfigure}[b]{0.32\textwidth}
        \includegraphics[width=\textwidth]{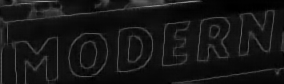}
        \caption{Reconstructed Edge Map}
    \end{subfigure}
    \hspace*{\fill}
	\begin{subfigure}[b]{0.32\textwidth}
        \includegraphics[width=\textwidth]{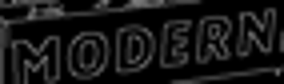}
        \caption{Bicubic upsampling}
    \end{subfigure}
    \hspace*{\fill}
    
    \caption{Canny edge map reconstruction}
\end{figure*}

\subsection{Loss Function}
Bicubic upsampled $C_{b}$ and $C_{r}$ channels of input LR are appended to the reconstructed Y channel. This is the residual HR image. Now, bicubic upsampled input LR image is added to it to form reconstructed HR image. Loss function is L2, which measures the mean squared pixel-wise difference between the ground truth and the reconstructed HR image.

\section{Experiments}
\subsection{Experimental Setup}
For all the experiments, we used Synth90k\cite{synth90k} dataset as the training data. It consists of 93K images. The color images were converted to $YC_{b}C_{r}$ and only Y-channel is processed. LR images are created by downsizing the images using bicubic kernel at both scales (2$\times$ and 4$\times$). 16x16 patches were generated and 20 patches are used as a mini-batch to train the model. This is the only pre-processing step.

We implemented the proposed architectures in the Tensorflow framework and all of the experiments were conducted on a GeForce GTX 1080ti GPU. We have used Adam\cite{adam} optimizer with momentum term 0.9. The initial learning rate is 2e-3 and it is halved after every epoch. Training is stopped once the learning rate reaches 2e-5.

\subsection{Evaluation}
We have examined the quality of character outline-edges in the reconstructed HR images, by passing canny edge maps of the text LR image as input to the model. Fig. 2 shows the sharpness of reconstructed edge maps with a scale of 4$\times$. The performance is extensively evaluated on eight popular benchmark datasets to cover various scenarios of scenic text. TABLE \RomanNumeralCaps{2}  draws a comparison between bicubic upsampling and our approach using the obtained PSNR and SSIM metrics for scale 2 and scale 4. These metrics have been calculated on the entire dataset, not just the on test set. These are listed as : ICDAR 2015 TextSR, ICDAR 2003 (IC03)\cite{ic03}, ICDAR 2013 (IC13)\cite{ic13}, ICDAR 2015 (IC15)\cite{ic15}, SVT\cite{svt}, IIIT5K\cite{5k}, CUTE80\cite{cute80}, MLT\footnote{https://rrc.cvc.uab.es/?ch=15}.

HTM Tran et al.\cite{hanh} have trained LapSRN\cite{lapsrn} with Gradient Difference Loss\cite{gdl}. They have proposed a new dataset Text330\footnote{https://github.com/t2mhanh/LapSRN\_TextImages\_git} which contains plain text document images. Apart from scenic text images, our model also performs well on document images. This is illustrated in TABLE \RomanNumeralCaps{1}. Fig.3 demonstrates sharper character edges in the reconstructed HR images from the Text330 test set as compared to bicubic upsampling. The given results confirm enhanced readability.

\begin{table}[htpb]
\centering
\caption{Evaluation on Test330 dataset}
\begin{tabular}[t]{ccccc}
\toprule
Method& Scale& PSNR& SSIM&\\
\toprule
Bicubic& 2& 21.82& 0.869& \\
LapSRN \cite{lapsrn}& 2& 26.65& 0.954& \\
LapSRN L1-GDL \cite{hanh}& 2& 30.10& 0.997& \\
Ours& 2& \textbf{31.06}& \textbf{0.997}&\\
\midrule
\\Bicubic& 4& 18.21& 0.692& \\
LapSRN \cite{lapsrn}& 4& 19.64& 0.796& \\
LapSRN L1-GDL \cite{hanh}& 4& 21.62& 0.875& \\
Ours& 4& \textbf{21.94}& \textbf{0.883}&\\
\bottomrule
\end{tabular}
\end{table}

SRCNN\cite{srcnn} is a three layer network having basic configuration $f_{1} = 9$, $f_{2} = 1$, $f_{3} = 5$, $n_{1} = 64$, $n_{2} = 32$, $n_{3} = 1$. It can be denoted as 64(9)-32(1)-1(5). 14-model combination\cite{dong} proposed by C Dong et al. achieved a PSNR of around 33 dB. Their best single model 64(9)-32(7)-16(5)-1(5) with 116,382 parameters achieved a PSNR of 31.99 dB. However, our model with 57,564 parameters achieves a PSNR of 31.38 dB on the same test. We also achieve an OCR accuracy of \textbf{75.89\%} on the same dataset using Tesseract\footnote{Tesseract version 3.02,http://code.google.com/p/tesseract-ocr}, whereas the original high-resolution images give \textbf{78.10\%} OCR accuracy. Fig. 4 and Fig. 5 show our result on LR images with different resolutions at 2$\times$ and 4$\times$ scales respectively. Our solution runs on the Samsung Galaxy S10 device with an average inference time of \textbf{11.7} milliseconds per image.

\section{Conclusion}
In this work, we have addressed the problem of enhancing text clarity in LR images, which is very common in mobile devices, and optimized our network for running on the edge. To the best of our knowledge, there are not many works trying to solve the problem of text readability with SR  methods on device. Our evaluation on PSNR, SSIM, on-device inference time and OCR accuracy shows that we have gained significant improvement compared to the existing Bicubic upsampling. We have also shown impressive results on document images. Unlike previous SR solutions, our proposed method focuses on SR of the text present in the image, rather than background or texture. This makes our model the very first lightweight model that is capable of text image SR in real-time on device.

\begin{table*}[htpb]
\caption{Evaluation on various benchmark datasets}
  \label{our-result-table}
  \begin{center}
  \begin{tabular}{cccccccccccc}
    \toprule
    \multirow{2}{*}{Method} &   
    \multirow{2}{*}{Scale} &
    \multirow{2}{*}{Metrics} & 
    \multicolumn{7}{c} {Datasets} \\
      & & & & {TextSR} &{IC13} & {IC15} & {IC03} & {SVT} & {IIIT5K} & {CUTE80} & {MLT}\\ 
    \toprule
   \multirow{2}{*} {Bicubic} & \multirow{2}{*} {2} & PSNR && 23.50 & 31.91 & 30.58 & 27.57 & 36.60 & 23.78 & 34.2 & 36.55 \\
    & & SSIM &&  0.879 & 0.9017 & 0.8983 & 0.8588 & 0.9570 &  0.8087& 0.9351 & 0.9579\\
\\     
     \multirow{2}{*} {Ours} & \multirow{2}{*} {2} & PSNR && 31.38 & 37.12 & 35.16 & 33.37 & 40.83&  29.25 & 39.5 & 42.76\\
      && SSIM && 0.9784 & 0.9432 & 0.9065 & 0.9083 & 0.9761 &  0.8872 & 0.9639 & 0.9846
\\
\\

	 \multirow{2}{*} {Ours (w/o edge features)} & \multirow{2}{*} {2} & PSNR && 31.32 & 37.05 & 35.12 & 33.33 & 40.75&  29.2 & 39.46 & 42.64\\
      && SSIM && 0.9712 & 0.9407 & 0.9058 & 0.9078 & 0.9740 &  0.8865 & 0.9617 & 0.9837\\
\midrule
\\      
 \multirow{2}{*} {Bicubic} & \multirow{2}{*} {4} & PSNR && - & 25.49 & 24.02 & 21.97 & 28.11 &  18.49 & 28.85 & 29.16\\
          && SSIM && - & 0.6902 & 0.5949 & 0.5913 & 0.7697 & 0.5936 & 0.8431 & 0.8438 \\
\\ 
 \multirow{2}{*} {Ours} & \multirow{2}{*} {4}  & PSNR && - & 29.94 & 26.47 & 26.39 & 32.05  & 21.22 & 33.49  & 36.63 \\
          && SSIM && - & 0.7922 & 0.6205 & 0.6823 & 0.8467 & 0.6878 & 0.9046 & 0.9507 \\
 \\
 \multirow{2}{*} {Ours (w/o edge features)} & \multirow{2}{*} {4}  & PSNR && - & 29.88 & 26.43 & 26.32 & 31.99  & 21.19 & 33.45  & 36.55 \\
          && SSIM && - & 0.7880 & 0.6185 & 0.6788 & 0.8467 & 0.6865 & 0.9001 & 0.9484 \\
\bottomrule
  \end{tabular}
\end{center}
\end{table*}


\begin{figure*}[t!]
    \hspace*{\fill}
    \begin{subfigure}[b]{0.32\textwidth}
        \includegraphics[width=\textwidth]{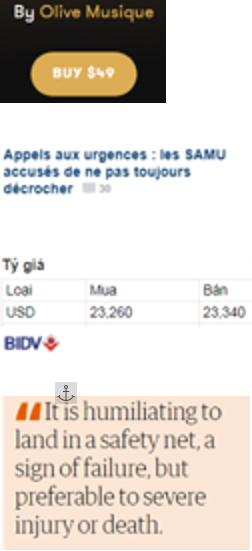}
        \caption{Bicubic}
    \end{subfigure}
    \hspace*{\fill}
    \begin{subfigure}[b]{0.32\textwidth}
        \includegraphics[width=\textwidth]{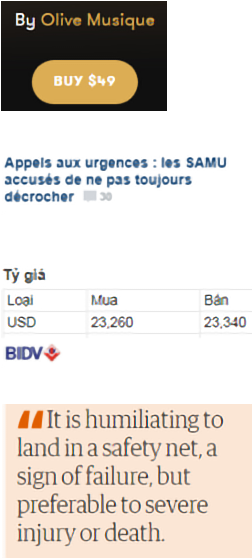}
        \caption{Our result}
    \end{subfigure}
    \hspace*{\fill}
	\begin{subfigure}[b]{0.32\textwidth}
        \includegraphics[width=\textwidth]{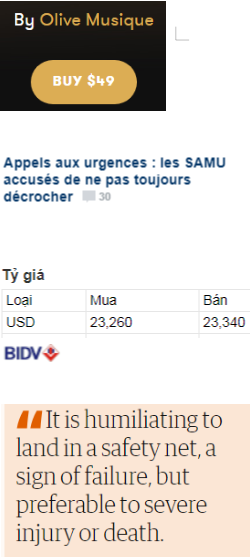}
        \caption{Ground Truth}
    \end{subfigure}
    \hspace*{\fill}
    \caption{Our results on Test330 dataset}
\end{figure*}

\newpage

\begin{figure*}[t!]
    \hspace*{\fill}
    \begin{subfigure}[b]{0.3\textwidth}
        \includegraphics[width=\textwidth]{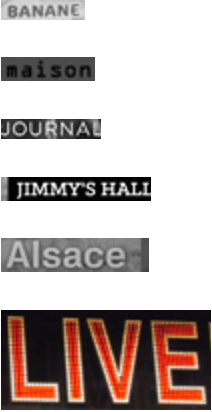}
        \caption{Bicubic}
    \end{subfigure}
    \hspace*{\fill}
    \begin{subfigure}[b]{0.3\textwidth}
        \includegraphics[width=\textwidth]{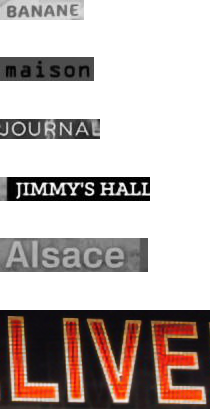}
        \caption{Our result}
    \end{subfigure}
    \hspace*{\fill}
	\begin{subfigure}[b]{0.3\textwidth}
        \includegraphics[width=\textwidth]{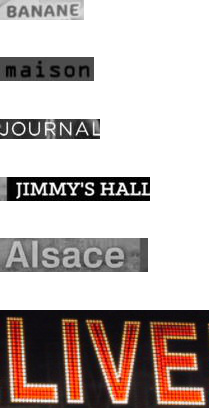}
        \caption{Ground Truth}
    \end{subfigure}
    \hspace*{\fill}
    \caption{Our results on scale 2}
\end{figure*}

\begin{figure*}[t!]
    \hspace*{\fill}
    \begin{subfigure}[b]{0.3\textwidth}
        \includegraphics[width=\textwidth]{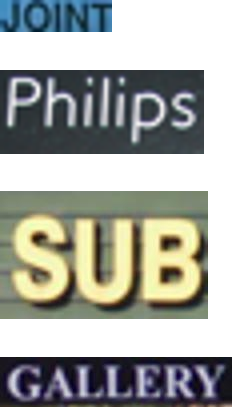}
        \caption{Bicubic}
    \end{subfigure}
    \hspace*{\fill}
    \begin{subfigure}[b]{0.3\textwidth}
        \includegraphics[width=\textwidth]{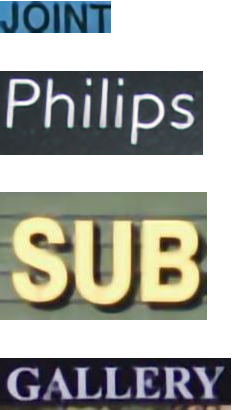}
        \caption{Our result}
    \end{subfigure}
    \hspace*{\fill}
	\begin{subfigure}[b]{0.3\textwidth}
        \includegraphics[width=\textwidth]{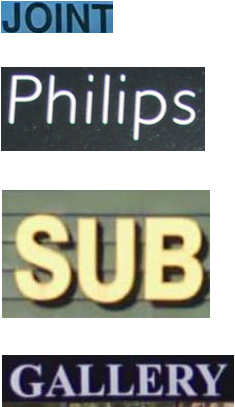}
        \caption{Ground Truth}
    \end{subfigure}
    \hspace*{\fill}
    \caption{Our results on scale 4}
\end{figure*}

\bibliographystyle{IEEEtran}
\bibliography{IEEEabrv,references}

\begin{thebibliography}{10}
\providecommand{\url}[1]{#1}
\csname url@samestyle\endcsname
\providecommand{\newblock}{\relax}
\providecommand{\bibinfo}[2]{#2}
\providecommand{\BIBentrySTDinterwordspacing}{\spaceskip=0pt\relax}
\providecommand{\BIBentryALTinterwordstretchfactor}{4}
\providecommand{\BIBentryALTinterwordspacing}{\spaceskip=\fontdimen2\font plus
\BIBentryALTinterwordstretchfactor\fontdimen3\font minus
  \fontdimen4\font\relax}
\providecommand{\BIBforeignlanguage}[2]{{%
\expandafter\ifx\csname l@#1\endcsname\relax
\typeout{** WARNING: IEEEtran.bst: No hyphenation pattern has been}%
\typeout{** loaded for the language `#1'. Using the pattern for}%
\typeout{** the default language instead.}%
\else
\language=\csname l@#1\endcsname
\fi
#2}}
\providecommand{\BIBdecl}{\relax}
\BIBdecl

\bibitem{allebach1996edge}
J.~Allebach and P.~W. Wong, ``Edge-directed interpolation,'' in
  \emph{Proceedings of 3rd IEEE International Conference on Image Processing},
  vol.~3.\hskip 1em plus 0.5em minus 0.4em\relax IEEE, 1996, pp. 707--710.

\bibitem{li2001new}
X.~Li and M.~T. Orchard, ``New edge-directed interpolation,'' \emph{IEEE
  transactions on image processing}, vol.~10, no.~10, pp. 1521--1527, 2001.

\bibitem{zhang2006edge}
L.~Zhang and X.~Wu, ``An edge-guided image interpolation algorithm via
  directional filtering and data fusion,'' \emph{IEEE transactions on Image
  Processing}, vol.~15, no.~8, pp. 2226--2238, 2006.

\bibitem{sun2010gradient}
J.~Sun, Z.~Xu, and H.-Y. Shum, ``Gradient profile prior and its applications in
  image super-resolution and enhancement,'' \emph{IEEE Transactions on Image
  Processing}, vol.~20, no.~6, pp. 1529--1542, 2010.

\bibitem{chang2004super}
H.~Chang, D.-Y. Yeung, and Y.~Xiong, ``Super-resolution through neighbor
  embedding,'' in \emph{Proceedings of the 2004 IEEE Computer Society
  Conference on Computer Vision and Pattern Recognition, 2004. CVPR 2004.},
  vol.~1.\hskip 1em plus 0.5em minus 0.4em\relax IEEE, 2004, pp. I--I.

\bibitem{gao2012image}
X.~Gao, K.~Zhang, D.~Tao, and X.~Li, ``Image super-resolution with sparse
  neighbor embedding,'' \emph{IEEE Transactions on Image Processing}, vol.~21,
  no.~7, pp. 3194--3205, 2012.

\bibitem{zhu2014single}
Y.~Zhu, Y.~Zhang, and A.~L. Yuille, ``Single image super-resolution using
  deformable patches,'' in \emph{Proceedings of the IEEE Conference on Computer
  Vision and Pattern Recognition}, 2014, pp. 2917--2924.

\bibitem{yang2008image}
J.~Yang, J.~Wright, T.~Huang, and Y.~Ma, ``Image super-resolution as sparse
  representation of raw image patches,'' in \emph{2008 IEEE conference on
  computer vision and pattern recognition}.\hskip 1em plus 0.5em minus
  0.4em\relax IEEE, 2008, pp. 1--8.

\bibitem{yang2013fast}
J.~Yang, Z.~Lin, and S.~Cohen, ``Fast image super-resolution based on in-place
  example regression,'' in \emph{Proceedings of the IEEE conference on computer
  vision and pattern recognition}, 2013, pp. 1059--1066.

\bibitem{yang2012coupled}
J.~Yang, Z.~Wang, Z.~Lin, S.~Cohen, and T.~Huang, ``Coupled dictionary training
  for image super-resolution,'' \emph{IEEE transactions on image processing},
  vol.~21, no.~8, pp. 3467--3478, 2012.

\bibitem{dong2014learning}
C.~Dong, C.~C. Loy, K.~He, and X.~Tang, ``Learning a deep convolutional network
  for image super-resolution,'' in \emph{European conference on computer
  vision}.\hskip 1em plus 0.5em minus 0.4em\relax Springer, 2014, pp. 184--199.

\bibitem{dong2016accelerating}
C.~Dong, C.~C. Loy, and X.~Tang, ``Accelerating the super-resolution
  convolutional neural network,'' in \emph{European conference on computer
  vision}.\hskip 1em plus 0.5em minus 0.4em\relax Springer, 2016, pp. 391--407.

\bibitem{vdsr}
J.~Kim, J.~Kwon~Lee, and K.~Mu~Lee, ``Accurate image super-resolution using
  very deep convolutional networks,'' in \emph{Proceedings of the IEEE
  conference on computer vision and pattern recognition}, 2016, pp. 1646--1654.

\bibitem{drcn}
------, ``Deeply-recursive convolutional network for image super-resolution,''
  in \emph{Proceedings of the IEEE conference on computer vision and pattern
  recognition}, 2016, pp. 1637--1645.

\bibitem{resnet}
K.~He, X.~Zhang, S.~Ren, and J.~Sun, ``Deep residual learning for image
  recognition,'' in \emph{Proceedings of the IEEE conference on computer vision
  and pattern recognition}, 2016, pp. 770--778.

\bibitem{mao2016image}
X.~Mao, C.~Shen, and Y.-B. Yang, ``Image restoration using very deep
  convolutional encoder-decoder networks with symmetric skip connections,'' in
  \emph{Advances in neural information processing systems}, 2016, pp.
  2802--2810.

\bibitem{cui2014deep}
Z.~Cui, H.~Chang, S.~Shan, B.~Zhong, and X.~Chen, ``Deep network cascade for
  image super-resolution,'' in \emph{European Conference on Computer
  Vision}.\hskip 1em plus 0.5em minus 0.4em\relax Springer, 2014, pp. 49--64.

\bibitem{goodfellow2014generative}
I.~Goodfellow, J.~Pouget-Abadie, M.~Mirza, B.~Xu, D.~Warde-Farley, S.~Ozair,
  A.~Courville, and Y.~Bengio, ``Generative adversarial nets,'' in
  \emph{Advances in neural information processing systems}, 2014, pp.
  2672--2680.

\bibitem{srgan}
C.~Ledig, L.~Theis, F.~Husz{\'a}r, J.~Caballero, A.~Cunningham, A.~Acosta,
  A.~Aitken, A.~Tejani, J.~Totz, Z.~Wang \emph{et~al.}, ``Photo-realistic
  single image super-resolution using a generative adversarial network,'' in
  \emph{Proceedings of the IEEE conference on computer vision and pattern
  recognition}, 2017, pp. 4681--4690.

\bibitem{capel}
D.~Capel and A.~Zisserman, ``Super-resolution enhancement of text image
  sequences,'' in \emph{Proceedings 15th International Conference on Pattern
  Recognition. ICPR-2000}, vol.~1.\hskip 1em plus 0.5em minus 0.4em\relax IEEE,
  2000, pp. 600--605.

\bibitem{nasonov}
A.~V. Nasonov and A.~S. Krylov, ``Text images superresolution and
  enhancement,'' in \emph{2012 5th International Congress on Image and Signal
  Processing}.\hskip 1em plus 0.5em minus 0.4em\relax IEEE, 2012, pp. 617--620.

\bibitem{btv}
S.~Farsiu, M.~D. Robinson, M.~Elad, and P.~Milanfar, ``Fast and robust
  multiframe super resolution,'' \emph{IEEE transactions on image processing},
  vol.~13, no.~10, pp. 1327--1344, 2004.

\bibitem{banerjee}
J.~Banerjee and C.~Jawahar, ``Super-resolution of text images using
  edge-directed tangent field,'' in \emph{2008 The Eighth IAPR International
  Workshop on Document Analysis Systems}.\hskip 1em plus 0.5em minus
  0.4em\relax IEEE, 2008, pp. 76--83.

\bibitem{lat}
A.~Lat and C.~Jawahar, ``Enhancing ocr accuracy with super resolution,'' in
  \emph{2018 24th International Conference on Pattern Recognition
  (ICPR)}.\hskip 1em plus 0.5em minus 0.4em\relax IEEE, 2018, pp. 3162--3167.

\bibitem{pandey2018binary}
R.~K. Pandey, K.~Vignesh, A.~Ramakrishnan \emph{et~al.}, ``Binary document
  image super resolution for improved readability and ocr performance,''
  \emph{arXiv preprint arXiv:1812.02475}, 2018.

\bibitem{sredgenet}
C.~Wang, Y.~Liu, X.~Bai, W.~Tang, P.~Lei, and J.~Zhou, ``Deep residual
  convolutional neural network for hyperspectral image super-resolution,'' in
  \emph{International Conference on Image and Graphics}.\hskip 1em plus 0.5em
  minus 0.4em\relax Springer, 2017, pp. 370--380.

\bibitem{edsr}
B.~Lim, S.~Son, H.~Kim, S.~Nah, and K.~Mu~Lee, ``Enhanced deep residual
  networks for single image super-resolution,'' in \emph{Proceedings of the
  IEEE conference on computer vision and pattern recognition workshops}, 2017,
  pp. 136--144.

\bibitem{dong}
C.~Dong, X.~Zhu, Y.~Deng, C.~C. Loy, and Y.~Qiao, ``Boosting optical character
  recognition: A super-resolution approach,'' \emph{arXiv preprint
  arXiv:1506.02211}, 2015.

\bibitem{combination}
W.~Ouyang, P.~Luo, X.~Zeng, S.~Qiu, Y.~Tian, H.~Li, S.~Yang, Z.~Wang, Y.~Xiong,
  C.~Qian \emph{et~al.}, ``Deepid-net: multi-stage and deformable deep
  convolutional neural networks for object detection,'' \emph{arXiv preprint
  arXiv:1409.3505}, 2014.

\bibitem{hanh}
H.~T. Tran and T.~Ho-Phuoc, ``Deep laplacian pyramid network for text images
  super-resolution,'' in \emph{2019 IEEE-RIVF International Conference on
  Computing and Communication Technologies (RIVF)}.\hskip 1em plus 0.5em minus
  0.4em\relax IEEE, 2019, pp. 1--6.

\bibitem{haochen}
H.~Zhang, D.~Liu, and Z.~Xiong, ``Cnn-based text image super-resolution
  tailored for ocr,'' in \emph{2017 IEEE Visual Communications and Image
  Processing (VCIP)}.\hskip 1em plus 0.5em minus 0.4em\relax IEEE, 2017, pp.
  1--4.

\bibitem{weighted}
G.~Seif and D.~Androutsos, ``Edge-based loss function for single image
  super-resolution,'' in \emph{2018 IEEE International Conference on Acoustics,
  Speech and Signal Processing (ICASSP)}.\hskip 1em plus 0.5em minus
  0.4em\relax IEEE, 2018, pp. 1468--1472.

\bibitem{lapsrn}
W.-S. Lai, J.-B. Huang, N.~Ahuja, and M.-H. Yang, ``Deep laplacian pyramid
  networks for fast and accurate super-resolution,'' in \emph{Proceedings of
  the IEEE conference on computer vision and pattern recognition}, 2017, pp.
  624--632.

\bibitem{mobisr}
R.~Lee, S.~I. Venieris, L.~Dudziak, S.~Bhattacharya, and N.~D. Lane, ``Mobisr:
  Efficient on-device super-resolution through heterogeneous mobile
  processors,'' in \emph{The 25th Annual International Conference on Mobile
  Computing and Networking}, 2019, pp. 1--16.

\bibitem{kd}
Q.~Gao, Y.~Zhao, G.~Li, and T.~Tong, ``Image super-resolution using knowledge
  distillation,'' in \emph{Asian Conference on Computer Vision}.\hskip 1em plus
  0.5em minus 0.4em\relax Springer, 2018, pp. 527--541.

\bibitem{gdl}
M.~Mathieu, C.~Couprie, and Y.~LeCun, ``Deep multi-scale video prediction
  beyond mean square error,'' \emph{arXiv preprint arXiv:1511.05440}, 2015.

\bibitem{srcnn}
C.~Dong, C.~C. Loy, K.~He, and X.~Tang, ``Image super-resolution using deep
  convolutional networks,'' \emph{IEEE transactions on pattern analysis and
  machine intelligence}, vol.~38, no.~2, pp. 295--307, 2015.

\bibitem{drrn}
Y.~Tai, J.~Yang, and X.~Liu, ``Image super-resolution via deep recursive
  residual network,'' in \emph{Proceedings of the IEEE conference on computer
  vision and pattern recognition}, 2017, pp. 3147--3155.

\bibitem{skipconnSR}
T.~Tong, G.~Li, X.~Liu, and Q.~Gao, ``Image super-resolution using dense skip
  connections,'' in \emph{Proceedings of the IEEE International Conference on
  Computer Vision}, 2017, pp. 4799--4807.

\bibitem{dcscn}
J.~Yamanaka, S.~Kuwashima, and T.~Kurita, ``Fast and accurate image super
  resolution by deep cnn with skip connection and network in network,'' in
  \emph{International Conference on Neural Information Processing}.\hskip 1em
  plus 0.5em minus 0.4em\relax Springer, 2017, pp. 217--225.

\bibitem{checkerboard}
A.~Aitken, C.~Ledig, L.~Theis, J.~Caballero, Z.~Wang, and W.~Shi,
  ``Checkerboard artifact free sub-pixel convolution: A note on sub-pixel
  convolution, resize convolution and convolution resize,'' \emph{arXiv
  preprint arXiv:1707.02937}, 2017.

\bibitem{synth90k}
M.~Jaderberg, K.~Simonyan, A.~Vedaldi, and A.~Zisserman, ``Synthetic data and
  artificial neural networks for natural scene text recognition,'' in
  \emph{Workshop on Deep Learning, NIPS}, 2014.

\bibitem{adam}
D.~P. Kingma and J.~Ba, ``Adam: A method for stochastic optimization,''
  \emph{arXiv preprint arXiv:1412.6980}, 2014.

\bibitem{ic03}
S.~M. Lucas, A.~Panaretos, L.~Sosa, A.~Tang, S.~Wong, R.~Young, K.~Ashida,
  H.~Nagai, M.~Okamoto, H.~Yamamoto \emph{et~al.}, ``Icdar 2003 robust reading
  competitions: entries, results, and future directions,'' \emph{International
  Journal of Document Analysis and Recognition (IJDAR)}, vol.~7, no. 2-3, pp.
  105--122, 2005.

\bibitem{ic13}
D.~Karatzas, F.~Shafait, S.~Uchida, M.~Iwamura, L.~G. i~Bigorda, S.~R. Mestre,
  J.~Mas, D.~F. Mota, J.~A. Almazan, and L.~P. De~Las~Heras, ``Icdar 2013
  robust reading competition,'' in \emph{2013 12th International Conference on
  Document Analysis and Recognition}.\hskip 1em plus 0.5em minus 0.4em\relax
  IEEE, 2013, pp. 1484--1493.

\bibitem{ic15}
D.~Karatzas, L.~Gomez-Bigorda, A.~Nicolaou, S.~Ghosh, A.~Bagdanov, M.~Iwamura,
  J.~Matas, L.~Neumann, V.~R. Chandrasekhar, S.~Lu \emph{et~al.}, ``Icdar 2015
  competition on robust reading,'' in \emph{2015 13th International Conference
  on Document Analysis and Recognition (ICDAR)}.\hskip 1em plus 0.5em minus
  0.4em\relax IEEE, 2015, pp. 1156--1160.

\bibitem{svt}
K.~Wang, B.~Babenko, and S.~Belongie, ``End-to-end scene text recognition,'' in
  \emph{2011 International Conference on Computer Vision}.\hskip 1em plus 0.5em
  minus 0.4em\relax IEEE, 2011, pp. 1457--1464.

\bibitem{5k}
A.~Mishra, K.~Alahari, and C.~V. Jawahar, ``Scene text recognition using higher
  order language priors,'' in \emph{BMVC}, 2012.

\bibitem{cute80}
A.~Risnumawan, P.~Shivakumara, C.~S. Chan, and C.~L. Tan, ``A robust arbitrary
  text detection system for natural scene images,'' \emph{Expert Systems with
  Applications}, vol.~41, no.~18, pp. 8027--8048, 2014.

\end{thebibliography}

\end{document}